\documentclass[10pt,twocolumn,letterpaper]{article}
\usepackage[pagenumbers]{cvpr}  
\usepackage[pagebackref,breaklinks,colorlinks]{hyperref}
\usepackage{algorithm}
\usepackage{algorithmic}
\usepackage{makecell}
\usepackage{amssymb}
\usepackage{amsmath}
\usepackage{xr}
\usepackage{tikz}
\usepackage{pgfplots}
\usepackage{subcaption}

\usepackage[capitalize]{cleveref}
\crefname{section}{Sec.}{Secs.}
\Crefname{section}{Section}{Sections}
\Crefname{table}{Table}{Tables}
\crefname{table}{Tab.}{Tabs.}

\makeatletter
\DeclareRobustCommand\onedot{\futurelet\@let@token\@onedot}
\def\@onedot{\ifx\@let@token.\else.\null\fi\xspace}

\def\eg{\emph{e.g}\onedot} 
\def\ie{\emph{i.e}\onedot} 
 
\def\etc{\emph{etc}\onedot} 
 
\def\etal{\emph{et al}\onedot}
\makeatother

\definecolor{bblue}{HTML}{4F81BD}
\definecolor{rred}{HTML}{C0504D}
\definecolor{ggreen}{HTML}{9BBB59}
\definecolor{ppurple}{HTML}{9F4C7C}

\setcitestyle{square}

\def\confName{CVPR}
\def\confYear{2023}

\title{Factorized-Dreamer: Training A High-Quality Video Generator with Limited and Low-Quality Data}

\author{Tao Yang$^{1}$,  Yangming Shi$^{1}$,  Yunwen Huang$^{1}$,  Feng Chen$^{1}$, Yin Zheng$^{1}$, Lei Zhang$^{2}$ \\
{$^{1}$ByteDance Inc.} \qquad $^{2}$The Hong Kong Polytechnic University
}

\usepackage{bibentry}

\begin{document}

\maketitle

\begin{abstract}
Text-to-video (T2V) generation has gained significant attention due to its wide applications to video generation, editing, enhancement and translation, \etc. However, high-quality (HQ) video synthesis is extremely challenging because of the diverse and complex motions existed in real world. Most existing works struggle to address this problem by collecting large-scale HQ videos, which are inaccessible to the community. In this work, we show that publicly available limited and low-quality (LQ) data are sufficient to train a HQ video generator without recaptioning or finetuning. We factorize the whole T2V generation process into two steps: generating an image conditioned on a highly descriptive caption, and  synthesizing the video conditioned on the generated image and a concise caption of motion details. Specifically, we present \emph{Factorized-Dreamer}, a factorized spatiotemporal framework with several critical designs for T2V generation, including an adapter to combine text and image embeddings, a pixel-aware cross attention module to capture pixel-level image information, a T5 text encoder to better understand motion description, and a PredictNet to supervise optical flows. We further present a noise schedule, which plays a key role in ensuring the quality and stability of video generation. Our model lowers the requirements in detailed captions and HQ videos, and can be directly trained on limited LQ datasets with noisy and brief captions such as WebVid-10M, largely alleviating the cost to collect large-scale HQ video-text pairs. Extensive experiments in a variety of T2V and image-to-video generation tasks demonstrate the effectiveness of our proposed Factorized-Dreamer. Our source codes are available at \url{https://github.com/yangxy/Factorized-Dreamer/}.
\end{abstract}

%

\section{Introduction}

Video generation is a highly challenging task due to the difficulties in simulating the complex and dynamic real-world scenarios conditioned on a text prompt, a label or an image. While in computer graphics, we can resort to physical laws to simulate fluid \cite{yang2015fluid}, cloth \cite{narain2012cloth}, \etc, such methods are limited in specific objects and scenes. Video generation has recently achieved impressive progress with the rapid development of deep learning techniques \cite{skorokhodov2021stylegan_v,singer2022makeavideo,blattmann2023svd}. Early attempts \cite{skorokhodov2021stylegan_v,brooks2022longvideo} are mostly based on generative adversarial networks (GANs) \cite{goodfellow2014gan}. Though yielding promising results, GAN-based video generation models suffer from unstable performance and restricted scenarios. With the emerging of diffusion models \cite{ho2020ddpm}, large-scale text-to-image (T2I) models \cite{rombach2021latent,ramesh2022dalle2,saharia2022imagen} trained on web-scale data can be used to generate high-quality images with a wide range of diversity and aesthetics. This encourages the research on training text-to-video (T2V) models \cite{singer2022makeavideo,ho2022imagenvideo,blattmann2023svd}. While much progress has been made, video generation still lags behind image generation in terms of  quality and diversity as it needs to model an extra temporal dimension, which involves of complex changes of scene content and object motions/actions. 

\begin{figure*}[t!]
    \centering
    \includegraphics[width=0.96\textwidth]{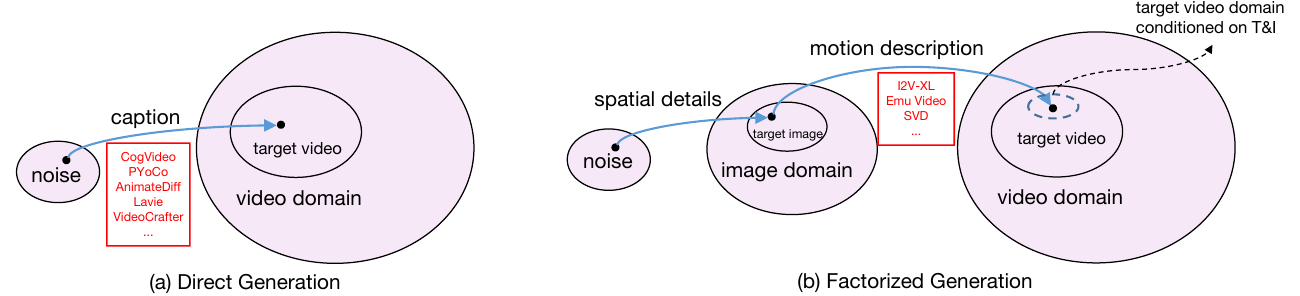}
    \vspace*{-1mm}
    \caption{Illustration of common video generation frameworks.}
    \label{fig:motivation}
\end{figure*}

The research on video generation can be divided into two categories: direct generation and factorized generation, as depicted in Fig.~\ref{fig:motivation}. The former learns a direct mapping from a text prompt to a generated video, while the latter resorts to an intermediate image generated by an off-the-shelf T2I model, and focuses on synthesizing a video conditioned on both the image and the text prompts. Video generation has been significantly boosted by diffusion models \cite{ho2020ddpm,rombach2021latent}. Ho \etal \cite{ho2022video} trained the first video diffusion model from scratch. Following works \cite{singer2022makeavideo,khachatryan2023text2video-zero,guo2023animatediff,blattmann2023svd} are mostly built on pre-trained T2I models such as Stable Diffusion (SD) \cite{rombach2021latent} and PixArt-$\alpha$ \cite{chen2023pixartalpha} considering the fact that a video is composed of a sequence of frames. By using pre-trained T2I models to generate spatial structures, these methods focus on how to encode motion dynamics into the latent codes \cite{khachatryan2023text2video-zero} or insert additional temporal layers \cite{singer2022makeavideo,guo2023animatediff,blattmann2023svd} to model the temporal dimension of videos. Along this line of research, many methods have been proposed to further boost the generation performance via re-designing video noise prior \cite{ge2023pyoco}, inserting motion module into a personalized T2I model \cite{guo2023animatediff}, investigating the influence of data selection \cite{blattmann2023svd}, adjusting noise schedules \cite{girdhar2023emuvideo}, improving the training scheme \cite{wang2023modelscope,chen2024videocrafter2}, and so on. Though much progress has been achieved, these methods either yield unpleasant results \cite{wang2023modelscope,guo2023animatediff} or rely on self-collected large-scale dataset of text-video pairs, which are inaccessible to the community \cite{ge2023pyoco,blattmann2023svd,girdhar2023emuvideo}. 
Recently, Peebles and Xie \cite{peebles2022dit} introduced the diffusion transformer (DiT) for image generation, which has been adopted in state-of-the-art T2V \cite{ma2024latte,openai2024sora} model training. In particular, SORA \cite{openai2024sora} demonstrates very impressive ability to generate extremely realistic videos. However, the details of its architecture and training strategy are still under cover.

In this work, we investigate whether a high-quality (HQ) video generator can be trained using merely publicly available video datasets, even without recaptioning or finetuning. We follow the factorized generation framework and propose \emph{Factorized-Dreamer} to achieve this goal. We divide the T2V generation process into a T2I step and a subsequent text/image-to-video (TI2V) step. In this way, we alleviate the demand of accurate and descriptive video captions as the video synthesis is conditioned on the image which carries the needed spatial information. Meanwhile, the effect of low-quality (LQ) image features of video dataset such as WebVid-10M \cite{bain2021webvid} and their annoying watermark can be reduced in synthesizing HQ video frames.
Compared with existing factorized models \cite{zhang2023i2vgen,girdhar2023emuvideo,blattmann2023svd}, our proposed Factorized-Dreamer simplifies the training procedure by abandoning finetuning, and it is easier to train as it only needs to predict how the starting image evolves conditioned on the motion prompt, and can better retain the visual diversity, style, and quality of the off-the-shelf T2I model than direct T2V models \cite{guo2023animatediff,wang2023lavie,chen2024videocrafter2}.

More specifically, Factorized-Dreamer employs a  factorized spatiotemporal architecture. The spatial module inherits from the pretrained T2I model but uses a T2I adapter \cite{mou2023t2i-adapter} to produce extended text-image embeddings. The temporal module employs the pixel-aware cross attention (PACA) \cite{yang2023pasd} to perceive pixel-level information, and employs an LLM (\ie T5 \cite{raffel2020t5}) as the text encoder for better motion understanding. We also present a PredictNet to supervise the optical flow during training to boost the motion coherence for realistic video generation. 
Considering that the widely used noise schedules designed for T2I models are no longer suitable for T2V models, we present an adjusted noise schedule to ensure the quality and stability of video generation.  Extensive experiments demonstrate the effectiveness and flexibility of the proposed Factorized-Dreamer. Trained on the publicly accessible WebVid-10M, our Factorized-Dreamer can produce comparable or even better results than many T2V models trained with larger and self-collected data.

\section{Related Work}

\textbf{Text-to-Image Generation}.
Diffusion models \cite{ho2020ddpm} have recently demonstrated powerful capability in generating HQ natural images, surpassing traditional generative models such as GANs \cite{goodfellow2014gan}, VAEs \cite{kingma2014vae} and Flows \cite{rezende2015flow}. While early image generation models are conditioned on labels \cite{brock2018biggan} or focused on a specific object/scenery such as face \cite{karras2019stylegan}, the seminal work of DALLE \cite{ramesh2021dalle,ramesh2022dalle2} showed that T2I generation is possible. Saharia \etal \cite{saharia2022imagen} proposed Imagen, a cascade diffusion model, to achieve photo-realistic T2I generation. Such models work on pixel space and require extensive computational resources to train and inference. Rombach \etal \cite{rombach2021latent} instead employed a VAE model \cite{kingma2014vae} to map the image from pixel space to latent space and operated score-matching therein. This work was extended to Stable Diffusion (SD) \cite{rombach2021latent}, the first open-source large-scale pretrained T2I model. SD demonstrated that T2I diffusion priors are more powerful than GAN priors in handling diverse natural images \cite{zhang2023controlnet,ruiz2023dreambooth,brooks2022instructpix2pix,yang2023pasd}.
It has served as a source of inspiration for numerous subsequent works of T2I synthesis \cite{saharia2022imagen,podell2023sdxl,chen2023pixartalpha,esser2024sd3}, conditional generation \cite{zhang2023controlnet}, personalized generation \cite{ruiz2023dreambooth}, image inpainting \cite{yang2022paintbyexample}, image editing \cite{brooks2022instructpix2pix}, image super-resolution \cite{wang2023stablesr,yang2023pasd}, and so on. 

\textbf{Text-to-Video Generation}.
T2V generation \cite{singer2022makeavideo,ho2022imagenvideo,blattmann2023svd} has gained increasing attention with the huge success of T2I synthesis \cite{ramesh2022dalle2,rombach2021latent,saharia2022imagen,chen2023pixartalpha}. Most prior works leveraged T2I models for T2V generation. Khachatryan \etal \cite{khachatryan2023text2video-zero} generated videos by encoding motion dynamics in the latent space of a T2I model such as SD \cite{rombach2021latent}. Wu \etal \cite{wu2023tuneavideo} targeted at one-shot T2V generation by finetuning a T2I model with a single video. Though these methods can produce videos without training on large-scale video datasets, they are limited in the quality and diversity of generated videos. Since a video is composed of a sequence of frames, many works instead embed temporal modules into a T2I model and learn a direct mapping from a text or an image prompt to a generated video \cite{singer2022makeavideo,ho2022imagenvideo,wang2023modelscope,blattmann2023svd,girdhar2023emuvideo}. For example, Singer \etal \cite{singer2022makeavideo} trained a T2V model based on a pretrained T2I model with a mixture of image and video
datasets. Imagen-Video \cite{ho2022imagenvideo} extends the capacity of Imagen \cite{saharia2022imagen} to video synthesis by employing a cascade diffusion models.

The aforementioned works are trained in pixel domain, suffering from the challenge of modeling high-dimensional spatiotemporal space. Many following works resort to latent diffusion models for video generation \cite{blattmann2023videoldm,ge2023pyoco,guo2023animatediff,blattmann2023svd,girdhar2023emuvideo}. Blattmann \etal \cite{blattmann2023videoldm} synthesized videos by merely training the newly added temporal layers. Ge \etal \cite{ge2023pyoco} improved the synthesis quality by introducing a video noise prior. Guo \etal \cite{guo2023animatediff} discovered that replacing the base T2V model with a personalized one could boost the performance. Based on \cite{blattmann2023videoldm}, Blattmann \etal \cite{blattmann2023svd} proposed Stable Video diffusion (SVD) with a comprehensive training and data selection strategy. Girdhar \etal \cite{girdhar2023emuvideo} explored new noise schedules for diffusion and a multi-stage training to achieve more stable results. Recently, a couple of T2V models \cite{ma2024latte,openai2024sora} emerged based on the DiT architecture \cite{peebles2022dit,chen2023pixartalpha}. In particular, SORA \cite{openai2024sora} exhibits impressive realistic long video generation performance. Though great progress has been made, previous methods either produce LQ results \cite{wang2023modelscope} or build on the proprietary datasets that are not available to public \cite{girdhar2023emuvideo,openai2024sora}. In this work, we argue that a reasonably high-quality video generator can be built on purely publicly available datasets without recaptioning \cite{betker2023dalle3} or finetuning. 

\textbf{Leverage Knowledge from Images for Video Generation}.
Large T2I models \cite{ramesh2022dalle2,rombach2021latent,saharia2022imagen,chen2023pixartalpha} are all trained on web-scale image-text pairs. It is supposed that T2V models require a significantly larger training dataset than T2I ones due to the extra temporal dimension. Unfortunately, publicly available video-text datasets are typically an order of magnitude smaller than image-text datasets \cite{bain2021webvid}, and they are limited in style and quality. To alleviate this problem, different strategies have been proposed to leverage knowledge from image data for video generation. For example, we can firstly train a T2I model and then finetune it partially \cite{blattmann2023videoldm,guo2023animatediff} or entirely \cite{wang2023modelscope} on the video dataset, train the T2V model jointly on image-text and video-text pairs from scratch \cite{ho2022video,singer2022makeavideo,ho2022imagenvideo}, concatenate the first frame as input \cite{blattmann2023svd,girdhar2023emuvideo}, and use CLIP
image features as a condition \cite{blattmann2023svd,zhang2023i2vgen}. In this work, we leverage the image knowledge by factorizing the T2V generation into two sub-problems: (1) generating an image using a text prompt based on an off-the-shelf T2I model; and (2) synthesizing a final video based on the image and text conditions.

\section{Method}
\subsection{Motivation}
When conditioned on the same text prompt, the target domain of a T2V model is significantly larger than that of T2I due to the extra temporal dimension (see Fig.~\ref{fig:motivation}). This means that a HQ T2V generator should be much bigger than a T2I one, and therefore a larger HQ training dataset is expected. Unfortunately, publicly available text-video datasets \cite{bain2021webvid} are typically of LQ and are significantly smaller than text-image datasets \cite{schuhmann2022laion5b}. Many existing T2V models \cite{singer2022makeavideo,blattmann2023svd} are built on pretrained T2I models with inserted temporal layers, and their parameters are basically on the same level of T2I models. We argue that these two issues block the training a satisfied T2V generator. 

To address the above issues, it is critical to shrink the target domain conditioned on the text prompt. One solution is to recaption \cite{betker2023dalle3} the dataset to obtain more descriptive captions because a more detailed and accurate caption would make the target space more specific. However, this requires additional work load, and it cannot address the LQ image features and watermark issues inherited from the LQ text-video datasets. We instead employ a factorized solution, as shown in Fig.~\ref{fig:motivation}(b), by leveraging image knowledge for video generation. In this way, recaptioning is no longer needed as the image input has carried all the necessary spatial information, and the text prompt could be brief and focused on motion description. In addition, this solution can alleviate the annoying LQ image features and watermark issues because the generated HQ image is used to generate video during inference. As a result, the proposed factorized framework makes training a HQ video generator using merely publicly available datasets possible.

\begin{figure*}[t!]
    \centering
    \includegraphics[width=0.96\textwidth,height=8cm]{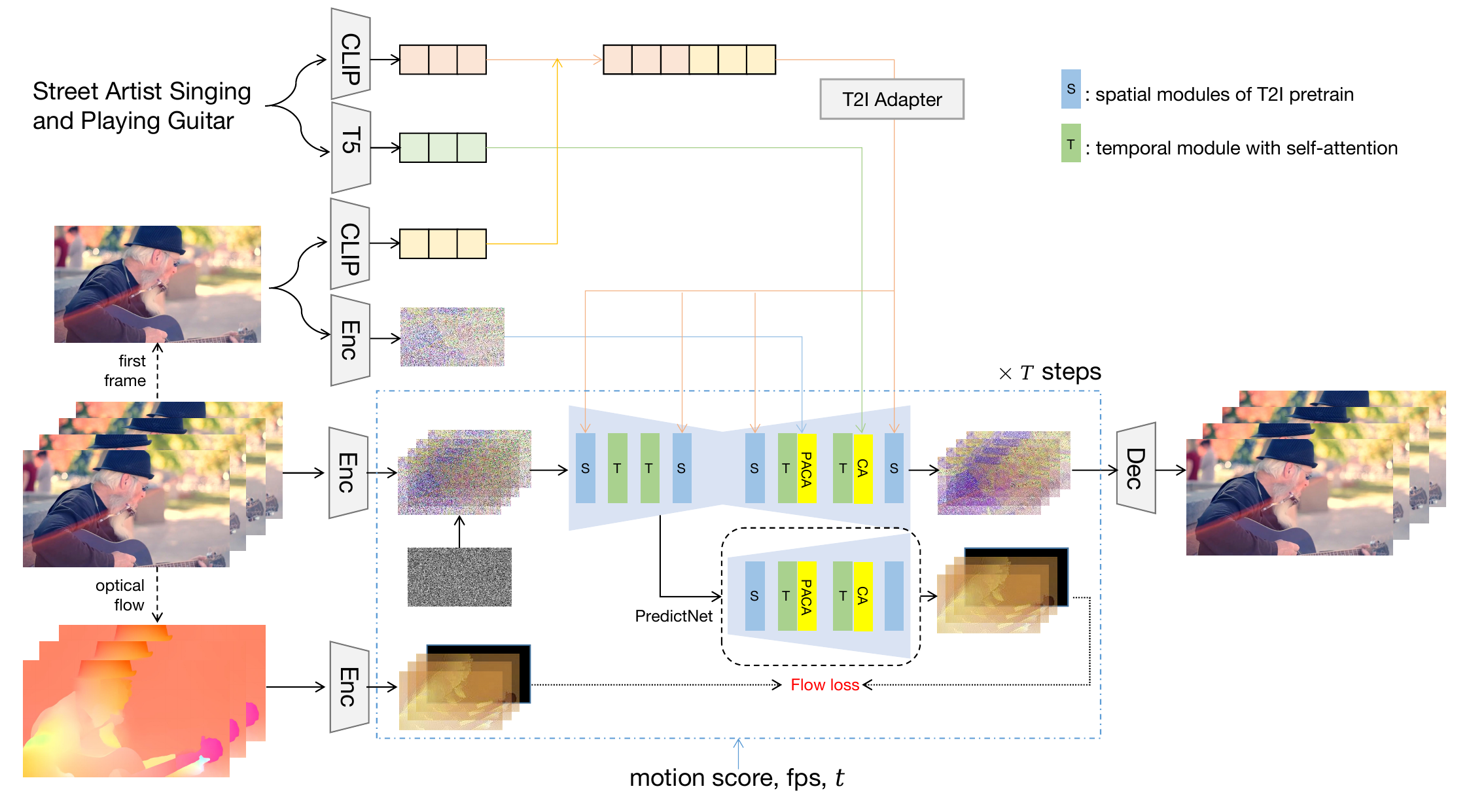}
    \vspace*{-2mm}
    \caption{Architecture of the proposed Factorized-Dreamer, which consists of a T2I adapter, PACA modules, a T5 text encoder and a PredictNet. During training, the encoder maps the input video to a latent representation, which is then added by noise. The noisy latent is fed to the UNet along with the latent first frame, the T5 text embedding, and a combined text-image embedding by the T2I adapter. The latent first frame and the combined text-image embedding are added to the UNet via PACA and CA in the upsampling layers, respectively. A PredictNet is introduced to enhance the motion coherence by supervising optical flows.}
    \label{fig:arch}
    \vspace*{-2mm}
\end{figure*}

\subsection{Architecture}
The architecture of the proposed Factorized-Dreamer is illustrated in Fig.~\ref{fig:arch}. By using an off-the-shelf T2I model to synthesize the first frame, the core part of Factorized-Dreamer lies in the TI2V model. Specifically, the proposed TI2V model consists of several factorized spatial and temporal modules with a few critical design decisions, including a T2I adapter to combine both image and text information for spatial modules, the pixel-aware cross attention (PACA) module for perceiving pixel-level image information, an LLM text encoder, \ie T5 \cite{raffel2020t5} for more precise motion understanding, and a PredictNet for supervising optical flows. 

\paragraph{Text-to-Image Adapter.}
\label{sec:adapter}
The captions in publicly available video datasets, \ie, WebVid \cite{bain2021webvid} and PexelVideos \cite{pexelvideo}, are usually precise but unfortunately concise. For instance, the caption shown in Fig.~\ref{fig:arch} accurately describes the subject and the activity in the video, but presents no detailed information about the man and the scenery. Without recaptioning the dataset, we instead use CLIP image features \cite{radford2021clip} as the conditional information with the help of a T2I adapter \cite{mou2023t2i-adapter}. 

Given a text and an image, we obtain the text and image embeddings $emb_{txt}, emb_{img}$ using CLIP \cite{radford2021clip}. In SD \cite{rombach2021latent}, the cross attention can be computed as follows:
\begin{equation}
\mathbf{Z}=CA(to\_q(\mathbf{x}), to\_kv(emb_{txt})),
\end{equation}
where $\mathbf{Z}, \mathbf{x}, to\_q, to\_kv$ are respectively the output of cross attention, the latent feature, a function that maps latent feature to query, and a function that maps $emb_{txt}$ to the key and the value. With the adapter, we update the computation of cross attention as follows:
\begin{align}
\mathbf{Z}=&CA(to\_q(\mathbf{x}), to\_kv(emb_{txt})) \nonumber \\
    &+ \lambda * CA(to\_q(\mathbf{x}), to\_kv\_adp(emb_{img})),
\end{align}
where $to\_kv\_adp$ maps the $emb_{img}$ to the key and the value, and $\lambda$ is a balancing parameter. We simply set $\lambda=1$ in our experiment. 

\paragraph{Pixel-Aware Cross Attention.}
\label{sec:paca}
The CLIP image features mainly provides global image information. To perceive pixel-level details, we resort to PACA \cite{yang2023pasd}, which is a plug-and-play module and has been successfully used in the task of image super-resolution \cite{yang2023pasd}. Unlike commonly used cross attention where the function $to\_kv$ takes CLIP embeddings as input, in PACA the input of $to\_kv$ is the conditioned image latent $\mathbf{y}$ encoded by the same VAE \cite{esser2021taming} of SD:
\begin{equation}
\mathbf{Z}=PACA(to\_q(\mathbf{x}), to\_kv(\mathbf{y})).
\end{equation}


\paragraph{Text Encoder for Motion Understanding.}
\label{sec:t5}
SD employs the CLIP text encoder \cite{radford2021clip} to align with the image embeddings. While trained on large-scale image-text pairs, the training data of CLIP are still significantly smaller than that used by LLMs \cite{raffel2020t5}. Furthermore, the captions in image-text pairs are mainly descriptions of image contents, which can be biased. Recent researches \cite{esser2024sd3} have showed that CLIP text encoder is less reliable in prompt following and typography
generation than LLMs \cite{raffel2020t5}. We argue that it also limits the motion understanding in video generation tasks. Therefore, we employ the widely used T5 \cite{raffel2020t5} to encode the caption for temporal modules. 

\paragraph{PredictNet for Motion Supervision.}
\label{sec:predictnet}
Motion coherence is one of the main challenges in video generation. Most existing works implicitly address this problem by resorting to large-scale video datasets \cite{blattmann2023svd}. Due to the limited sizes of publicly available video datasets, we introduce the \emph{PredictNet} to explicitly supervise the learning of video motions. The PredictNet is initialized by the upsampling part of the UNet, inspired by ControlNet \cite{zhang2023controlnet}. Unlike the UNet, it predicts the optical flows of input videos. Given a video latent $\mathbf{x}_0$, a randomly sampled diffusion step $t$ and various conditions $\mathbf{c}$, the PredictNet $p_\theta$ learns to predict the original optical flow $f$ using the $\mathbf{x}_0$-prediction:
\begin{equation}
\mathcal{L}_{flow}=\mathbb{E}_{\mathbf{x}_0,t,\mathbf{c},\epsilon\sim\mathcal{N}(0,1)}[||f-p_\theta(\mathbf{x}_t,t,\mathbf{c})||^2_2].
\end{equation}
PredictNet is only used in the final training stage. Our experiments demonstrate that the proposed PredictNet improves the coherence of generated videos. 

\subsection{Noise Schedule}
Diffusion models synthesize HQ data by progressively adding noise to a dataset and then learning to reverse this process, in which the noise schedule plays a key role. A number of works have studied the setting of noise schedule to improve the image \cite{lin2023snr,hoogeboom2023shift,chen2023scale} and video \cite{blattmann2023svd,menpace2024snapvideo} generation performance. 
As discovered by prior works \cite{lin2023snr,hoogeboom2023shift}, the noise schedule used in SD would cause the signal-to-noise-ratio (SNR) issue when directly applied to video generation models. The reason is twofold. First, the noise schedule leaves some residual signal even at the terminal diffusion timestep $N$ in training, leading to non-zero SNR. This weakens the model performance at test time due to train-test discrepancy. Second, given the noise with a certain intensity, applying it to video frames yields much stronger SNR than applying it to an image. This is because a clean image can be easily restored by simply averaging video frames. To address these issues, Lin \etal \cite{lin2023snr} re-scaled the noise schedule to enforce zero terminal SNR. Chen \cite{chen2023scale} scaled the input data by a factor. Hoogeboom \etal \cite{hoogeboom2023shift} introduced a shifted diffusion noise. Blattmann \etal \cite{blattmann2023svd} and Menpace \etal \cite{menpace2024snapvideo} introduced the EDM \cite{Karras2022edm} framework. 

For simplicity, we utilize the methods proposed by \cite{lin2023snr} and \cite{hoogeboom2023shift}. Given a video input with a resolution of $T\times H\times W$, its SNR at time $t\in\{1, 2...N\}$ can be updated as follows:
\begin{equation}
SNR^{T\times H\times W}(t)=SNR(t)*s^2,
\label{eqn:shift}
\end{equation}
where $SNR(t)=\frac{\bar{\alpha}_t}{1-\bar{\alpha}_t}$ is the original SNR in SD. $s=\sqrt{\frac{D\times D}{T\times H\times W}}$ is a shifting factor, where $D=256$ is the reference resolution. Moreover, we follow the idea proposed in \cite{lin2023snr} to re-scale the noise schedule as follows:
\begin{equation}
\sqrt{\bar{\alpha}'_t}=\frac{\sqrt{\bar{\alpha}_t-\bar{\alpha}_N}}{\sqrt{\bar{\alpha}_1-\bar{\alpha}_N}},
\label{eqn:rescale}
\end{equation}
where $\alpha'$ is the scaled $\alpha$. One can see that $\sqrt{\alpha'_{N-1}}\neq0$ and $\sqrt{\alpha'_N}=0$, which means that $\alpha'_N$ has been successfully set to $0$. We visualize the curves of log SNR to timestep $t$ in Fig.~\ref{fig:snr}. The shifting factor $s$ is $0.125$. One can see that after shifting and re-scaling the original SD noise schedule, the signal can be largely destroyed with more noise, making it more suitable for video generation tasks. 

\subsection{Training Strategy}
\label{sec:strategy}
In training, we learn the proposed Factorized-Dreamer model $\epsilon_\theta$ to predict the noise added to the noisy video latent $\mathbf{x}_t$ conditioned on $\mathbf{c}$. The optimization objective is:
\begin{equation}
\mathcal{L}_{DF}=\mathbb{E}_{\mathbf{x}_0,t,\mathbf{c},\epsilon\sim\mathcal{N}(0,1)}[||\epsilon-\epsilon_\theta(\mathbf{x}_t,t,\mathbf{c})||^2_2].
\end{equation}
In the final training stage, we jointly update the PredictNet. The total loss is $\mathcal{L}=\mathcal{L}_{DF}+\gamma\mathcal{L}_{flow}$, where we simply set the balancing parameter as $\gamma=1$. All parameters except those from CLIP, T5 and VAE are trainable. 

\section{Experiments}

\subsection{Experiment Setup}
\label{sec:exp}

\paragraph{Training and Testing Datasets.}
Unlike many previous methods \cite{blattmann2023svd,girdhar2023emuvideo} that are built on the proprietary data, we train our model on two publicly available datasets: (1) WebVid \cite{bain2021webvid} and an extension of it \cite{cleanvid} crawled from ShutterStock \footnote{https://www.shutterstock.com}, which contribute about $19M$ LQ text-video pairs after de-duplicating; and (2) PexelVideos \cite{pexelvideo}, which consists of about $300K$ HQ text-video pairs. Different from previous methods \cite{zhang2023i2vgen,chen2024videocrafter2,blattmann2023svd} that employ HQ data for an extra finetuning stage, we simply mixed the LQ and HQ datasets during the whole training procedure. We evaluate our model using the zero-shot T2V generation on the UCF101 dataset \cite{soomro2012ucf101}. 

\paragraph{Training Scheme.}
We use the AdamW \cite{kingma2015adam} optimizer with a fixed learning rate $5e^{-5}$ and a total batch size of $40$ videos. We adopt a multi-stage training strategy by firstly training on low-resolution (LR) videos, \ie, $32\times256\times256$, for $400k$ steps, then training on high-resolution (HR) video, \ie $16\times512\times512$, for $300k$ steps, and finally introducing the PredictNet to supervise motion flows for another $100k$ steps on the same HR videos in the second stage. The model is trained for about $12$ days with $40$ NVIDIA Tesla 80G-A100 GPUs.

\paragraph{Evaluation Metrics.}
For quantitative evaluation of our model on the T2V task, we exploit EvalCrafter \cite{liu2023evalcrafter}, which is a benchmark for evaluating T2V generation models \cite{chen2024videocrafter2}. It provides about $700$ prompts to conclude an objective metric from $4$ subjective studies, \ie, motion quality, text-video alignment, temporal consistency, and visual quality. We also employ the commonly used Frechet Video Distance (FVD) \cite{unterthiner2018fvd} and Inception Score (IS) \cite{salimans2016is} for zero-shot evaluation on the UCF101 \cite{soomro2012ucf101}. 
As for the I2V task, we employ Frame consistency (FC) and Prompt consistency (PC) as measures following previous methods \cite{girdhar2023emuvideo}. In addition, we conduct user studies for evaluation due to the lack of a comprehensive objective metric for T2V task. 

\begin{figure*}[t!]
    \centering
    \includegraphics[width=0.92\textwidth,height=8.5cm]{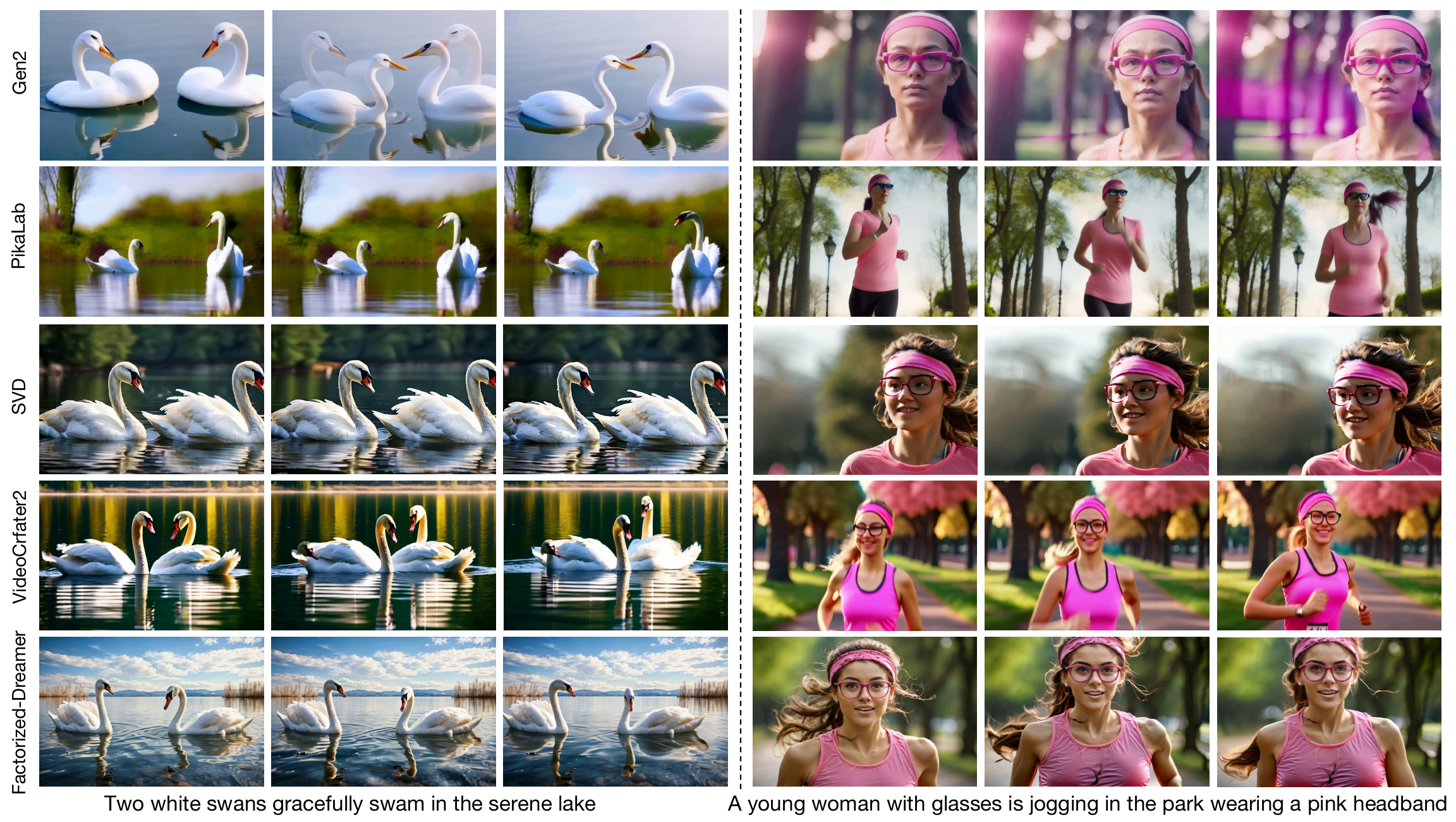}
    \vspace*{-3mm}
    \caption{Text-to-Video results by different methods.}
    \label{fig:t2v}
    \vspace*{-3mm}
\end{figure*}

\begin{table*}
    \centering
    \caption{Comparison of different T2V models on the EvalCrafter benchmark \protect{\cite{liu2023evalcrafter}}}
    \vspace*{-1mm}
    \normalsize{
    \resizebox{0.9\textwidth}{!}{\begin{tabular}{c|c c c c c|c c|c}
        Metrics & AnimateDiff & Lavie & VideoCrafter2 & PikaLab & Gen2 & I2V-XL & SVD & Factorized-Dreamer \\
      \hline\hline
        {Visual Quality$\uparrow$} & 61.17 & 57.99 & 63.98 & 63.05 & 69.09 & 53.09 & 63.74 & 67.12 \\
        {Text-Video Alignment$\uparrow$} & 69.25 & 68.49 & 63.16 & 66.97 & 63.92 & 54.46 & 57.08 & 65.02 \\
        {Motion Quality$\uparrow$} & 52.17 & 52.83 & 54.82 & 56.43 & 55.59 & 52.47 & 52.53 & 55.17 \\
        {Temporal Consistency$\uparrow$} & 62.34 & 54.23 & 61.46 & 63.81 & 65.4 & 57.8 & 58.70 & 63.43 \\
        {Sum Score$\uparrow$} & 245 & 234 & 243 & 250 & 254 & 218 & 232 & 251 \\
    \end{tabular}}}
    \label{tab:evalcrafter}
    \vspace*{-3mm}
\end{table*}

\begin{figure*}[t!]
    \centering
    \includegraphics[width=0.92\textwidth,height=10cm]{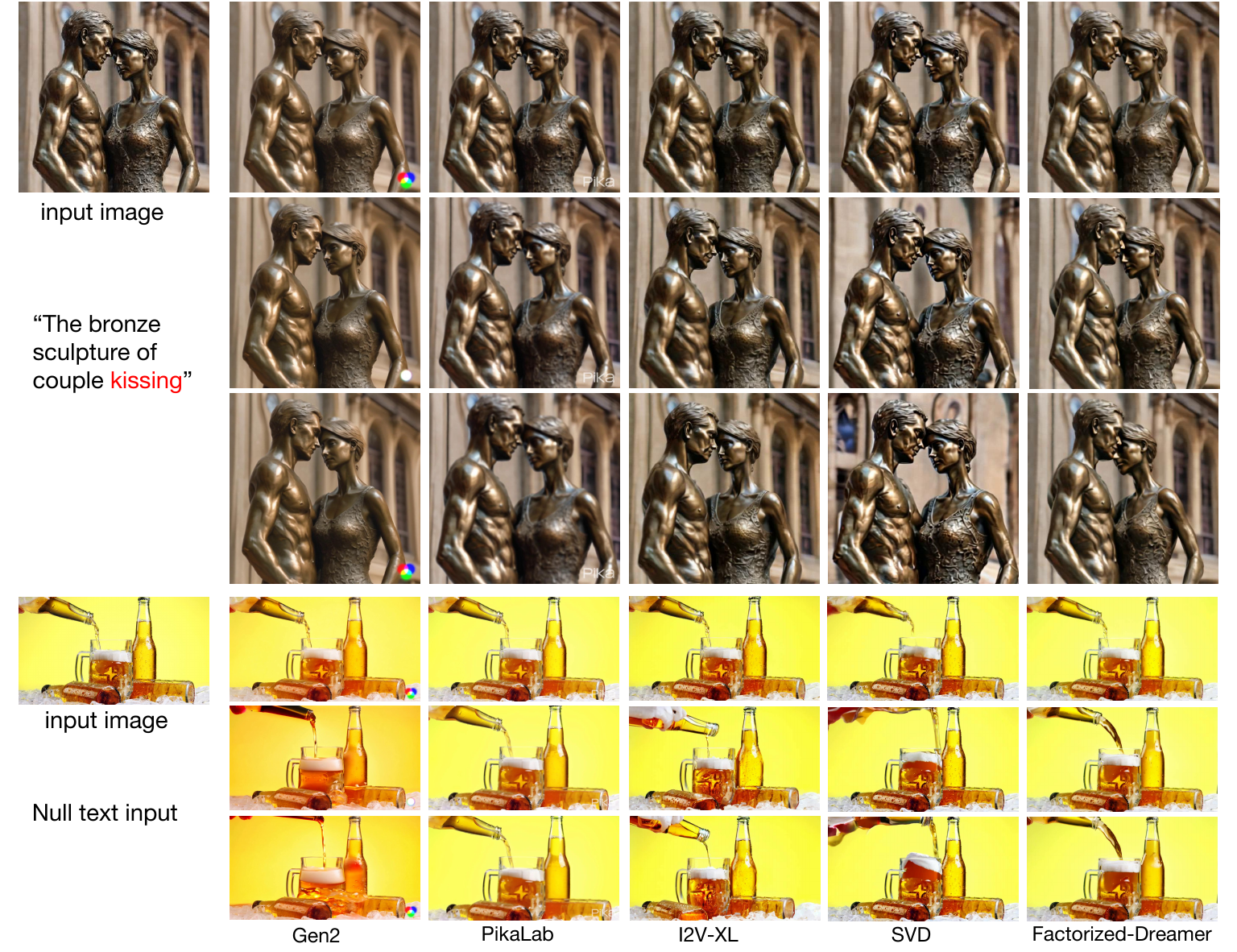}
    \vspace*{-3mm}
    \caption{Image-to-Video results by different methods.}
    \label{fig:i2v}
    \vspace*{-3mm}
\end{figure*}

\begin{table*}[t!]
    \begin{minipage}[b]{.45\textwidth}
    \centering
    \includegraphics[width=0.9\textwidth,height=5.5cm]{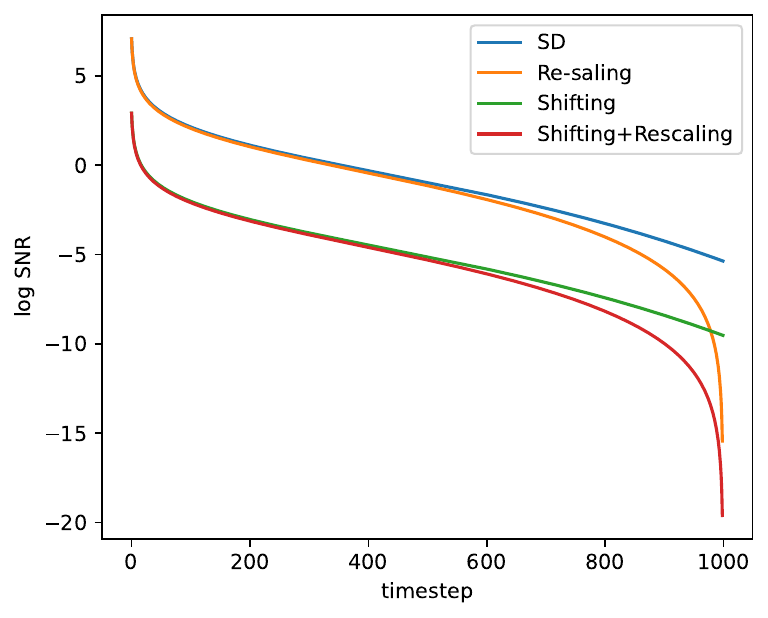}
    \vspace*{-3mm}
    \captionof{figure}{Curves of log SNR to timestep. $s=0.125$.} 
    \label{fig:snr}
    \vspace*{-1mm}
    \end{minipage}
    \hspace*{1mm}
    \begin{minipage}[b]{.45\textwidth}
    \centering
    \normalsize{
    \resizebox{0.9\textwidth}{!}{
    \begin{tabular}{c|c c}
      Method & FVD$\downarrow$ & IS$\uparrow$ \\
      \hline
      CogVideo(Chinese) \protect{\cite{hong2022cogvideo}} & 751.3 & 23.55 \\
      CogVideo(English) \protect{\cite{hong2022cogvideo}} & 701.6 & 25.27 \\
      MagicVideo \protect{\cite{zhou2022magicvideo}} & 655 & - \\
      Video LDM \protect{\cite{blattmann2023videoldm}} & 550.6 & 33.45 \\
      Make-A-Video \protect{\cite{singer2022makeavideo}} & 367.2 & 33.00 \\
      PYoCo \protect{\cite{ge2023pyoco}} & 355.2 & 47.46 \\
      Lavie \protect{\cite{wang2023lavie}} & 526.30 & - \\
      AnimateDiff \protect{\cite{guo2023animatediff}} & 598.83 & 35.18 \\
      I2V-XL \protect{\cite{zhang2023i2vgen}} & 424.87 & 28.78 \\
      VideoCrafter2 \protect{\cite{chen2024videocrafter2}} & 674.09 & 40.28 \\
      \hline
      Factorized-Dreamer & 503.93 & 33.27 \\
    \end{tabular}}}
    \caption{Zero-shot results on UCF101 \protect{\cite{soomro2012ucf101}}.} 
    \label{fig:zeroshot}
  \end{minipage}
  \vspace*{-2mm}
\end{table*}

\begin{table}[h!]
    \centering
    \normalsize{
    \resizebox{0.45\textwidth}{!}{\begin{tabular}{c|c c}
        Method & FC$\uparrow$ & PC$\uparrow$ \\
      \hline\hline
        I2V-XL \protect{\cite{zhang2023i2vgen}} & 0.9335 & 0.2723 \\
        SVD \protect{\cite{blattmann2023svd}} & 0.9736 & 0.2981 \\
        PikaLab \protect{\cite{pika}} & 0.9917 & 0.3261 \\
        Gen2 \protect{\cite{runway}} & 0.9897 & 0.3539 \\
        \hline
        Factorized-Dreamer & 0.9839 & 0.3500 \\
    \end{tabular}}}
    \vspace*{-2mm}
    \caption{The FC, PC indices of different I2V models.}
    \label{tab:fcpc}
    \vspace*{-4mm}
\end{table}

\begin{figure}[h!]
    \begin{subfigure}[b]{.45\textwidth}
    \centering
    \begin{tikzpicture}[scale=0.8]
    \begin{axis}[
        major x tick style = transparent,
        ybar=\pgflinewidth,
        bar width=16pt,
        ymajorgrids = true,
        ylabel = {User Preference},
        symbolic x coords={Visual Quality,Motion Quality},
        xtick = data,
        enlarge x limits=0.5,
        ymin=0,
        ylabel near ticks, yticklabel pos=left,
        legend cell align=left,
        legend style={
                at={(0.5,1.1)},
                anchor=north,
                column sep=1ex,
                legend columns=2
        }
    ]
        \addplot[style={bblue,fill=bblue,mark=none}]
            coordinates {(Visual Quality, 0.22) (Motion Quality,0.25)};

        \addplot[style={rred,fill=rred,mark=none}]
             coordinates {(Visual Quality,0.19) (Motion Quality,0.22)};

        \addplot[style={ggreen,fill=ggreen,mark=none}]
             coordinates {(Visual Quality,0.29) (Motion Quality,0.27) };

        \addplot[style={ppurple,fill=ppurple,mark=none}]
             coordinates {(Visual Quality,0.30) (Motion Quality,0.26)};

        \legend{Gen2,VideoCrafter2,PikaLab,Factorized-Dreamer}
    \end{axis}
    \end{tikzpicture}
    \caption{}
    \end{subfigure}
    \hspace{10mm}
    \begin{subfigure}[b]{.45\textwidth}
    \centering
    \begin{tikzpicture}[scale=0.9]
    \begin{axis}[
        major x tick style = transparent,
        ybar=\pgflinewidth,
        bar width=16pt,
        ymajorgrids = true,
        ylabel = {User Preference},
        symbolic x coords={Prompt Following,Motion Quality},
        xtick = data,
        enlarge x limits=0.5,
        ymin=0,
        ylabel near ticks, yticklabel pos=left,
        legend cell align=left,
        legend style={
                at={(0.5,1.1)},
                anchor=north,
                column sep=1ex,
                legend columns=2
        }
    ]
        \addplot[style={bblue,fill=bblue,mark=none}]
            coordinates {(Prompt Following, 0.3) (Motion Quality,0.25)};

        \addplot[style={rred,fill=rred,mark=none}]
             coordinates {(Prompt Following,0.15) (Motion Quality,0.26)};

        \addplot[style={ggreen,fill=ggreen,mark=none}]
             coordinates {(Prompt Following,0.28) (Motion Quality,0.26) };

        \addplot[style={ppurple,fill=ppurple,mark=none}]
             coordinates {(Prompt Following,0.27) (Motion Quality,0.23)};

        \legend{Gen2,SVD,PikaLab,Factorized-Dreamer}
    \end{axis}
    \end{tikzpicture}
    \caption{}
    \end{subfigure}
    \caption{User study results of (a) text-to-video generation and (b) image-to-video generation tasks.}
    \vspace*{-6mm}
\label{fig:ustudy}
\end{figure}
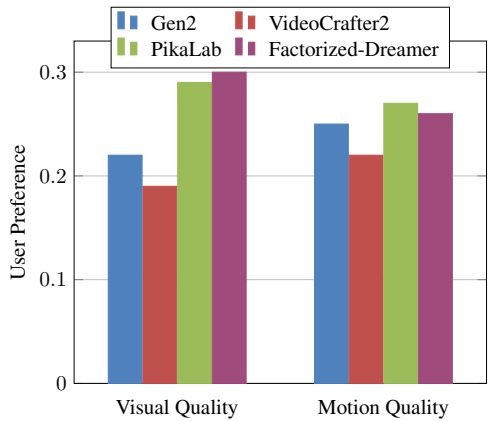
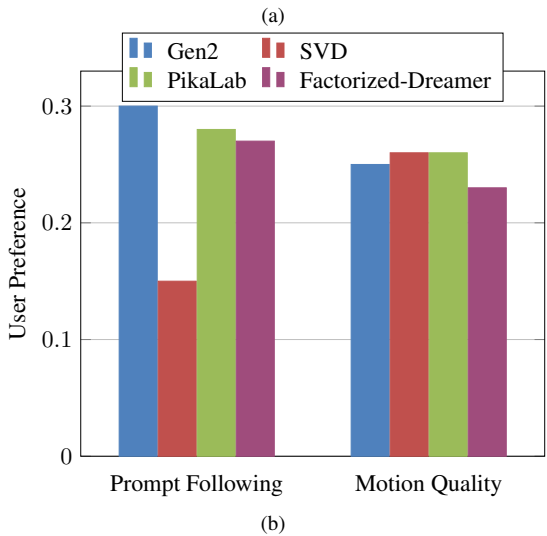

\begin{figure*}[t!]
    \centering
    \includegraphics[width=\textwidth]{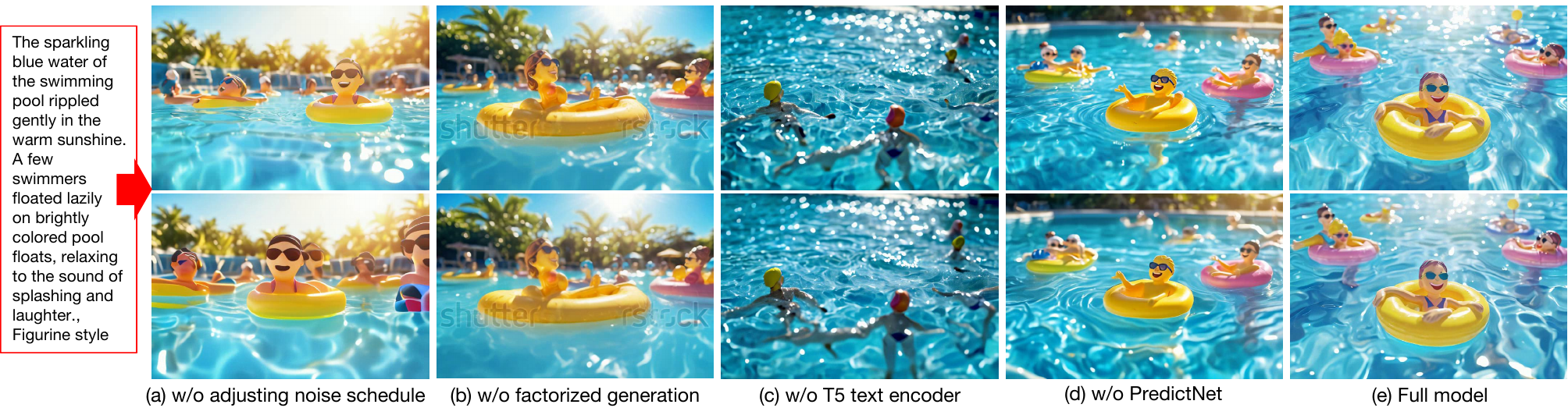}
    \vspace*{-5mm}
    \caption{Text-to-Video results by different variants of Factorized-Dreamer.}
    \label{fig:ablation}
    \vspace*{-2mm}
\end{figure*}

\begin{table*}[t!]
    \caption{Quantitative results of different variants of Factorized-Dreamer on EvalCrafter \protect{\cite{liu2023evalcrafter}}.}
    \centering
    \vspace*{-2mm}
    \normalsize{
    \resizebox{0.96\textwidth}{!}{\begin{tabular}{l|c c c c c|c c c c}
        Exp. & \makecell{Shifting \\ Noise Schedule} & \makecell{Rescaling \\ Noise Schedule} & \makecell{Factorized \\ Generation} & \makecell{T5 text \\ encoder} & PredictNet & \makecell{Visual \\ Quality}$\uparrow$ & \makecell{Text-Video \\ Alignment}$\uparrow$ & \makecell{Motion \\ Quality}$\uparrow$ & \makecell{Temporal \\ Consistency}$\uparrow$ \\
        \hline
        (a) & & & \checkmark & \checkmark & \checkmark & 54.45 & 52.79 & 51.07 & 55.84 \\
        (a*) & \checkmark & & \checkmark & \checkmark & \checkmark & 67.37 & 65.18 & 53.93 & 63.21 \\
        (b) & \checkmark & \checkmark & & \checkmark & \checkmark & 55.38 & 59.77 & 53.82 & 61.02 \\
        (c) & \checkmark & \checkmark & \checkmark & & \checkmark & 67.01 & 62.36 & 56.20 & 62.69 \\
        (d) & \checkmark & \checkmark & \checkmark & \checkmark & & 67.55 & 63.90 & 53.23 & 59.81 \\
        (e) & \checkmark & \checkmark & \checkmark & \checkmark & \checkmark & 67.12 & 65.02 & 55.17 & 63.43 \\
    \end{tabular}}}
    \vspace*{-3mm}
\label{tab:ablation}
\end{table*}

\subsection{Experimental Results}

\paragraph{Text-to-Video Generation.} 
We compare our proposed Factorized-Dreamer with two categories of video generation algorithms. The first category is the direct T2V generation models, including open-sourced AnimateDiff \cite{guo2023animatediff}, Lavie \cite{wang2023lavie} and VideoCrafter2 \cite{chen2024videocrafter2}, as well as the commercial Gen2 \cite{runway} and PikaLab \cite{pika}. The second category is factorized video generation models, including I2V-XL \cite{zhang2023i2vgen} and SVD \cite{blattmann2023svd}, where we use SDXL \cite{podell2023sdxl} in the T2I step for fair comparison.

We first compare against previous methods using the zero-shot T2V generation setting on UCF101 \cite{soomro2012ucf101}. As presented in Tab.~\ref{fig:zeroshot}, our Factorized-Dreamer achieves competitive FVD \cite{unterthiner2018fvd} and IS \cite{salimans2016is} scores. Since these metrics may not align with human preferences \cite{ho2022imagenvideo,girdhar2023emuvideo} due to the bias in training data, we further resort to the EvalCrafter benchmark \cite{liu2023evalcrafter}. The quantitative evaluation results are presented in Tab.~\ref{tab:evalcrafter}. 
On can see that our method achieves competitive visual quality score with PikaLab \cite{pika} and Gen2 \cite{runway} that are trained on large-scale HQ proprietary datasets, demonstrating the effectiveness of our factorized generation framework. Our Factorized-Dreamer ranks the second on motion quality and temporal consistency. This indicates the importance of our designs on the noise schedule and the PredictNet. Furthermore, the final sum score of Factorized-Dreamer surpasses all the open-source methods and only seconds to commercial Gen2 \cite{runway}. To conclude, Factorized-Dreamer confirms that training a HQ video generator using merely publicly available datasets is possible.

Fig.~\ref{fig:t2v} visualizes the T2V results of competing methods. It can be seen that the proposed Factorized-Dreamer can generate realistic videos with better visual quality. Interestingly, the synthesized videos are extremely coherent (\eg, even the movement of swan reflections are consistent), showing its potential of simulating physical world. More visual results can be found in the \textbf{supplementary material}.

\paragraph{Image-to-Video Generation.}
Our factorized generator can be directly used to animate a user-provided image. We compare Factorized-Dreamer with open-sourced I2V-XL \cite{zhang2023i2vgen}, SVD \cite{blattmann2023svd}, and the commercial Gen2 \cite{runway} and PikaLab \cite{pika}. All methods are conditioned on the same testing dataset, consisting of $20$ generated images by an off-the-shelf T2I model and $20$ collected images from the Internet. We report the evaluation results on FC and PC indices in Tab.~\ref{tab:fcpc}. On can see that our Factorized-Dreamer achieves superior performance than open-sourced methods. It slightly lags behind commercial approaches, which however use much more training data than us.

Fig.~\ref{fig:i2v} shows some I2V results of competing methods conditioning on a valid text prompt and a null text, respectively. It can be seen that the proposed Factorized-Dreamer can generate more text-aligned (see the kissing couple) and more physically realistic results (see the rising beer and foams in the cup during pouring). More visual results can be found in the \textbf{supplementary material}.


\subsection{User Studies}
We invite $15$ volunteers to conduct a user study for T2V and I2V tasks. Each volunteer is asked to choose the most preferred one among the outputs of all competing methods, which are presented to the volunteers in random order. We randomly select $15$ text prompts from EvalCrafter \cite{liu2023evalcrafter} and $15$ text-image pairs from AIGCBench \cite{fan2024aigcbench} for T2V and I2V, respectively. For T2V, we compare the proposed Factorized-Dreamer with Gen2 \cite{runway}, PikaLab \cite{pika}, and VideoCrafter2 \cite{chen2024videocrafter2} in terms of ``visual quality'' and ``motion quality'' axes. The axis of ``visual quality'' represents a comprehensive preference on both aesthetics and text-video alignment. As for I2V task, we compare our method with Gen2 \cite{runway}, PikaLab \cite{pika}, and SVD \cite{blattmann2023svd}. Since they are all conditioned on the same text and image prompts, we pay attention to their performance on ``prompt following'' and ``motion quality'' axes. Finally, we obtain $900$ votes, and the statistics are presented in Fig.~\ref{fig:ustudy}.

It can be seen that our Factorized-Dreamer outperforms competing methods on ``visual quality'' axis, and surpasses open-sourced method on ``prompt following''. For all tasks, our method achieves comparable ``motion quality'' performance to its competitors, even it is trained on merely publicly available limited and LQ datasets.

\subsection{Ablation Studies}
We perform a series of ablation experiments to study our design decisions, including the noise schedule, the importance of factorized generation, the role of T5 text encoder and PredictNet.

\paragraph{Effectiveness of Noise Schedule.}
We evaluate two variants of Factorized-Dreamer by adopting the original noise schedule of SD and the shifting operation in Eqn.~\ref{eqn:shift}. As shown in Fig.~\ref{fig:ablation}(a), the generation becomes highly unstable when directly employing the original SD noise schedule. This is also verified by the quantitative results of Exp. (a) in Tab.~\ref{tab:ablation}. Removing the rescaling operation in Eqn.~\ref{eqn:rescale} has little effect on the results (see evaluation results of Exp.(a*) in Tab.~\ref{tab:ablation}). This is because the residual signal at the terminal diffusion timestep has been largely reduced after shifting as shown in Fig.~\ref{fig:snr}, significantly alleviating the train-test discrepancy. 

\paragraph{Importance of Factorized Generation.}
To evaluate the importance of factorized generation, we remove the PACA module and the adapter module during model training. As a result, we obtain a direct T2V variant. As shown in Fig.~\ref{fig:ablation}(b) and Tab.~\ref{tab:ablation}, the direct T2V variant tends to yield inconsistent videos with low visual quality (\eg watermark) and worse quantitative indices (see Exps. (b) and (e)). 

\paragraph{Role of T5 Text Encoder.}
We test a variant by replacing T5 with the CLIP text encoder. It is found that this leads to occasional misunderstanding of motions. For example, the text prompt in Fig.~\ref{fig:ablation} emphasizes ``floating'' rather than ``swimming''. This is also confirmed by the reduced text-video alignment score in Tab.~\ref{tab:ablation} (see Exps. (c) and (e)).

\paragraph{Role of PredictNet.}
To verify how PredictNet benefits video generation, we conduct a variant by abandoning the jointly training stage as discussed in Sec.~\ref{sec:exp}. By comparing Exps. (d) and (e), it can be seen that the proposed PredictNet improves the motion quality and temporal consistency. In Fig.~\ref{fig:ablation}(d), one can see that an arm disappears during evolving without using PredictNet.

\section{Conclusion and Limitation}
We proposed a factorized framework, namely Factorized-Dreamer, for text based HQ video generation. Different from many existing methods that are trained on web-scale HQ proprietary datasets, we showed that a HQ video generator can be achieved by using merely publicly available, limited and LQ datasets, even without recaptioning or finetuning. We developed a factorized spatiotemporal architecture with several critical designs to accomplish this goal. The proposed Factorized-Dreamer was simple to implement, easy to train, and our extensive experiments demonstrated its effectiveness and flexibility across T2V and I2V tasks.

Though Factorized-Dreamer can synthesize HQ videos, it sometimes still suffers from inconsistent and incoherent motions due to limited training data. In addition, it struggles to generate long videos. A more sophistic framework like the one used in Sora could be  introduced to further improve the performance of Factorized-Dreamer, which will be considered in our future work.

{
    \small
    \bibliographystyle{ieeenat_fullname}
    \bibliography{main}
}

\end{document}